%% file: 2026_CVPR_JYJUN.tex
\documentclass[10pt,twocolumn,letterpaper]{article}

\usepackage[pagenumbers]{iccv} 

\input{preamble}

\usepackage{multirow}

\definecolor{cvprblue}{rgb}{0.21,0.49,0.74}
\usepackage[pagebackref,breaklinks,colorlinks,allcolors=cvprblue]{hyperref}

\newcommand{\bI}{\ensuremath{{\mathbf{I}}}}

\newcommand{\bx}{\ensuremath{{\mathbf{x}}}}
\newcommand{\bz}{\ensuremath{{\mathbf{z}}}}

\newcommand{\KGain}{\mathcal{K}_{\text{Gain}}}
\newcommand{\KeypointModel}{\mathcal{M}}

\newcommand{\FullDatasetName}{FidelityBench-258K}
\newcommand{\SmallDatasetName}{FidelityBench-300}

\usepackage{cuted}   
\usepackage{capt-of} 
\usepackage{etoolbox}
\makeatletter
\patchcmd{\@maketitle}{\vspace{12pt}}{\vspace{4pt}}{}{}
\makeatother

\usepackage[capitalize]{cleveref}
\crefname{section}{Sec.}{Secs.}
\Crefname{section}{Section}{Sections}
\Crefname{table}{Table}{Tables}
\crefname{table}{Tab.}{Tabs.}

\def\paperID{11903} 
\def\confName{CVPR}
\def\confYear{2026}

\title{FlowFixer: Towards Detail-Preserving Subject-Driven Generation}

\author{
Jinyoung Jun\textsuperscript{1,2}\thanks{Work done during internship at Amazon.} \quad
Won-Dong Jang\textsuperscript{1} \quad
Wenbin Ouyang\textsuperscript{1} \quad
Raghudeep Gadde\textsuperscript{1} \quad
Jungbeom Lee\textsuperscript{2}\thanks{Corresponding author.}\\[0.1em]
\textsuperscript{1}Amazon \quad
\textsuperscript{2}Korea University\\
{\tt\small {\{jyjun, wdjang, wenbinoy\}@amazon.com}} \quad
{\tt\small raghudeep.g@gmail.com} \quad
{\tt\small jbeomlee@korea.ac.kr} \\
}

\begin{document}
\maketitle
\vspace*{-50mm}

\begin{strip}
  \centering
  \includegraphics[width=\textwidth]{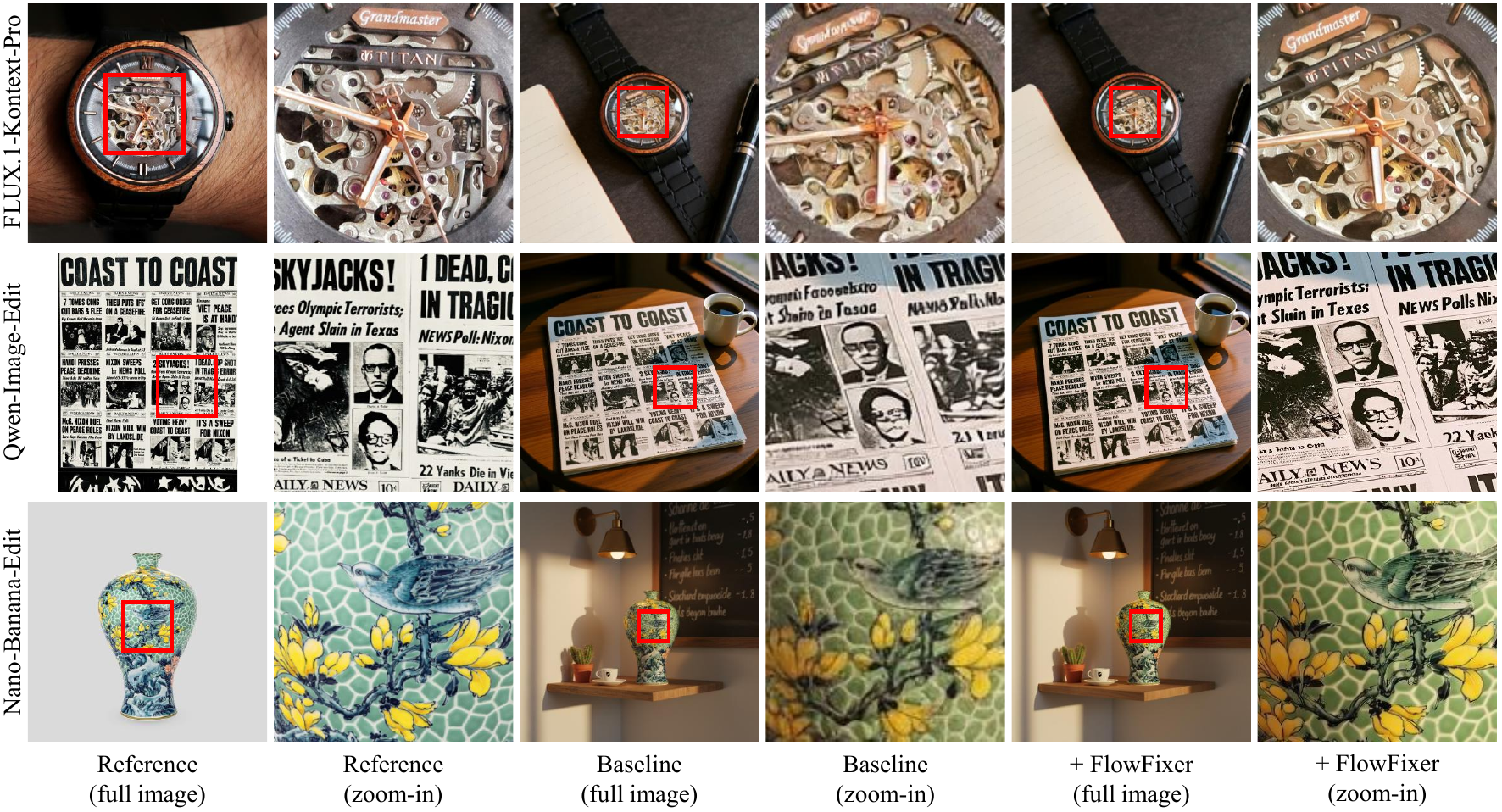}
  \vspace{-5mm}
  \captionof{figure}{\textbf{Detail enhancement on FLUX.1-Kontext-Pro~\cite{labs2025flux}, Qwen-Image-Edit~\cite{bai2023qwen}, and Nano-Banana-Edit~\cite{comanici2025gemini} using our FlowFixer.} The red boxes indicate the zoomed-in regions. 
  Compared to the baseline subject-driven generations, FlowFixer restores the fine details from the reference, such as complex structures (top and bottom), small text (top and middle), and human identity (middle). It also handles challenging cases involving rotation (top), viewpoint changes (middle), and color shifts (bottom), while preserving the overall scene composition.
  FlowFixer is a baseline-agnostic, prompt-free model designed to enhance subject fidelity without altering the global layout.}
  \label{fig:intro}
\end{strip}



\begin{abstract}
\vspace{-5mm} 

We present \textbf{FlowFixer}, a refinement framework for subject-driven generation (SDG) that restores fine details lost during generation caused by changes in scale and perspective of a subject. FlowFixer proposes direct image-to-image translation from visual references, avoiding ambiguities in language prompts. To enable image-to-image training, we introduce a one-step denoising scheme to generate self-supervised training data, which automatically removes high-frequency details while preserving global structure, effectively simulating real-world SDG errors. We further propose a keypoint matching-based metric to properly assess fidelity in details beyond semantic similarities usually measured by CLIP or DINO. Experimental results demonstrate that FlowFixer outperforms state-of-the-art SDG methods in both qualitative and quantitative evaluations, setting a new benchmark for high-fidelity subject-driven generation.


\end{abstract}
\vspace{-5mm}

\input{src/intro}
\input{src/related_work2}

\input{src/method}
\input{src/detail_aware_evaluation}

\input{src/experiments}

\input{src/conclusion}

\clearpage
{
    \small
    \bibliographystyle{ieeenat_fullname}
    \bibliography{2026_CVPR_JYJUN}
}


\end{document}

%% file: preamble.tex
\usepackage[dvipsnames]{xcolor}


%% file: src/intro.tex
\section{Introduction}
\label{sec:introduction}
Subject-driven generation (SDG) aims to embed a given subject (or an input reference image) into imagery described by an input text prompt while preserving the subject's identity. SDG has received significant attention from the community since it has a number of practical applications, including advertising content generation, short-form content generation, and personalized media creation.

Recent foundation models~\cite{peebles2023scalable, esser2024scaling} have shown promising improvements in handling subjects with simpler textures (\textit{e.g.}, animals or plain objects)~\cite{tan2025ominicontrol, tan2025ominicontrol2, wu2025less, xiao2025omnigen, wu2025omnigen2}. However, preserving complex product-specific details, such as logos, text, and intricate patterns, remains a critical challenge that demands greater attention from the community. This is particularly important for commercial applications where structural fidelity of product details directly impacts the utility of generated content. In advertising, for example, altered logos undermine brand recognition, and distorted text makes the outputs unusable.

There are two key obstacles underlying this difficulty. First, collecting high-quality paired training data for SDG is challenging. Ideally, one would need pairs of subject images and diverse ground truth images containing the same subject to supervise both fidelity and compositional diversity. In practice, however, collecting such data at scale is highly challenging. To address this scarcity, Subjects200K~\cite{tan2025ominicontrol} was introduced, yet it is constructed from synthetic images, which often lack fine-grained and realistic alignment of subject details.

Second, existing conditioning mechanisms are often limited in specifying fine-grained geometric and appearance variations of the subject. Text descriptions such as `a red sports car' or `a cereal box' convey only coarse appearance and provide limited cues about pose, orientation, or lighting, making precise reproduction of subject details challenging~\cite{li2023blip, ruiz2023dreambooth}. Even with image-based conditioning (e.g., depth or edge maps), they tend to prioritize global scene coherence over localized detail alignment, which can lead to the loss of high-frequency information in texture-rich or geometrically complex regions~\cite{zhang2023adding, tan2025ominicontrol, xiao2025omnigen}.

To overcome these challenges, we propose FlowFixer, a novel refinement framework for detail-preserving SDG. Our approach employs a direct image-to-image translation pipeline that learns from visual references. This design choice circumvents the ambiguity inherent in natural language descriptions, enabling precise preservation of diverse visual elements and fine structural details across the image, as illustrated in Figure~\ref{fig:intro}.

At the core of FlowFixer is a self-supervised refinement scheme that leverages pseudo-paired training data. In principle, training a subject refinement framework requires triplets consisting of a clean subject image, a corresponding SDG-generated image, and its ideal ground-truth for refinement. However, collecting such paired data at scale is impractical due to the high cost of annotating subject-scene correspondences and generating controlled SDG outputs.

Instead of collecting triplet data for training, we employ a self-supervised approach centered on our one-step denoising strategy. Starting with a clean real image, we synthetically generate its degraded counterpart through a forward diffusion step followed by single-step denoising using an off-the-shelf diffusion model. This process closely mimics SDG artifacts and characteristic distortions, allowing FlowFixer to learn fine detail restoration without expensive human supervision. The resulting framework enables efficient training using web-collected single images while faithfully representing the high-frequency detail loss typical in SDG applications.

Current quantitative evaluation metrics for SDG results have distinct characteristics and constraints. For example, pixel-level similarity measures (e.g., MSE or SSIM) focus primarily on low-level differences, while semantic-level metrics (e.g., FID~\cite{heusel2017gans} or LPIPS~\cite{zhang2018unreasonable}) may not fully capture fine details. Moreover, many existing metrics require ground truth images, which are often unavailable in real-world generative applications. To overcome these limitations, we propose detail-aware evaluation metrics based on keypoint matching~\cite{jiang2024omniglue, lindenberger2023lightglue}; absolute keypoint increase and keypoint matching gain. These metrics effectively capture structural fidelity by measuring the consistency between a reference image and its generated output, enabling ground-truth-free quantitative evaluation of detail preservation in open-world generative settings. Together with human and VLM evaluation, our metrics provide reliable and reproducible assessments of fine-grained fidelity.

\begin{figure}[t]
  \centering
  \includegraphics[width=\columnwidth]{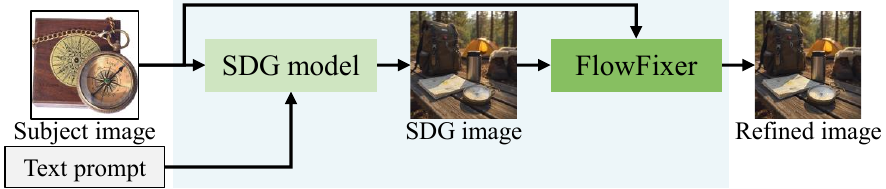}
  \caption{\textbf{FlowFixer overview.} FlowFixer enhances SDG images by restoring fine subject details, using the original subject image as reference.}
  \label{fig:overview_short}
\end{figure}

Through extensive experiments, we demonstrate that FlowFixer consistently outperforms existing SDG methods in preserving subject identity, establishing a new baseline for high-fidelity SDG. The key contributions of this work are summarized as follows:
\begin{itemize}
\itemsep0em

\item A novel model-agnostic refinement framework, FlowFixer, that substantially enhances subject fidelity in SDG-generated images.
\item An efficient training data curation pipeline based on one-step denoising, which effectively simulates diffusion artifacts to generate high-quality pseudo-paired training data.
\item A direct visual translation approach that leverages reference images, enabling precise preservation of visual elements and fine-grained details while eliminating prompt-induced ambiguity.
\item A novel ground-truth-free evaluation metric to assess visual fidelity based on keypoint matching, which demonstrates FlowFixer's superior detail preservation capability compared to existing SDG methods.

\end{itemize}

%% file: src/related_work2.tex
\section{Related Work}
\label{sec:related_work}
Subject-driven generation has received significant attention from the community and builds directly on top of the success of text-to-image foundational diffusion models \cite{rombach2022high, flux2024, labs2025flux}.
While text-to-image models can generate high quality objects, subject driven generation requires faithful rendering of the ``subject" (i.e, preserving the identity in the subject image) in a variety of scenes.

Techniques for subject driven generation have broadly followed two main directions (a) fine-tuning-based and (b) encoder-based. 
Early approaches such as DreamBooth~\cite{ruiz2023dreambooth}, Textual Inversion~\cite{gal2023image}, and LoRA~\cite{hu2022lora} in Custom Diffusion~\cite{kumari2023multi} adapt pre-trained text-to-image diffusion models to specific subjects using only a few reference images (typically 3 to 5), achieving strong identity preservation but require expensive per-subject fine-tuning.
More recent works avoid per-subject finetuning and address the limitation by injecting the reference image of the subject through an encoder directly into the diffusion backbone. 
For example, IP-Adapter~\cite{ye2023ip} injects image-prompt features via decoupled cross-attention to enable subject conditioning without fine-tuning, while BLIP-Diffusion~\cite{li2023blip} learns a multi-modal subject encoder for tighter subject–prompt alignment. 
OminiControl~\cite{tan2025ominicontrol} further shows that a DiT backbone can encode references natively with minimal additional parameters.

However, these encoder-based approaches, while good at preserving high-level details, struggle to preserve the subject's fine structural details, rendering the synthesized images unusable for real-world applications.
In contrast, FlowFixer restores missing low-level details to ensure high-fidelity identity preservation. 
It employs a reference-guided diffusion refinement process that corrects structural inconsistencies in a generated SDG image by conditioning on the original subject image, as illustrated in Figure~\ref{fig:overview_short}. 
This ``last mile" refinement makes FlowFixer universally compatible, enhancing identity preservation for any upstream model.



Another line of work that is relevant for subject driven image generation is based on image editing, which focuses on modifying an existing image under additional conditions such as text prompts, reference images, or spatial masks~\cite{zhang2023adding, mou2024t2i}.
Existing methods generally fall into two categories: global and local editing.
Global editing techniques alter overall image appearance or semantic content through text-based manipulation or latent space interpolation~\cite{hertz2022prompt, kawar2023imagic, meng2021sdedit, kim2022diffusionclip, parmar2023zero, patashnik2023localizing, cao2023masactrl, brack2024ledits++, tumanyan2023plug}.
On the other hand, methods for local editing target specific spatial regions via mask-based selection\cite{lugmayr2022repaint}, blending~\cite{avrahami2022blended}, or exemplars~\cite{gu2024analogist}.
Although effective for coarse transformations, these approaches often fail to preserve fine structural details of the subject and typically require manual inputs such as masks or region-specific prompts~\cite{couairon2022diffedit,wang2023imagen,yu2023inpaint,zhuang2024task}.
Recent works further reveal that text-driven editors struggle with localized or fine-grained control due to ambiguous conditioning and conflicting attention dynamics~\cite{zhou2025fireedit,mou2024diffeditor,guo2024focus,thakral2025fine,qiao2024baret}.



In contrast, FlowFixer performs automatic, reference-guided refinement without requiring manual annotations or textual conditioning. By leveraging cross-image correspondences between the generated and reference images, FlowFixer restores degraded regions while preserving global scene coherence and sharp, subject-consistent details. To further encourage the model to focus on subjects, the proposed FlowFixer exploits automatic subject cropping based on keypoint matching between a subject image and its SDG image.

%% file: src/method.tex
\begin{figure*}[t]
  \centering
  \includegraphics[width=\linewidth]{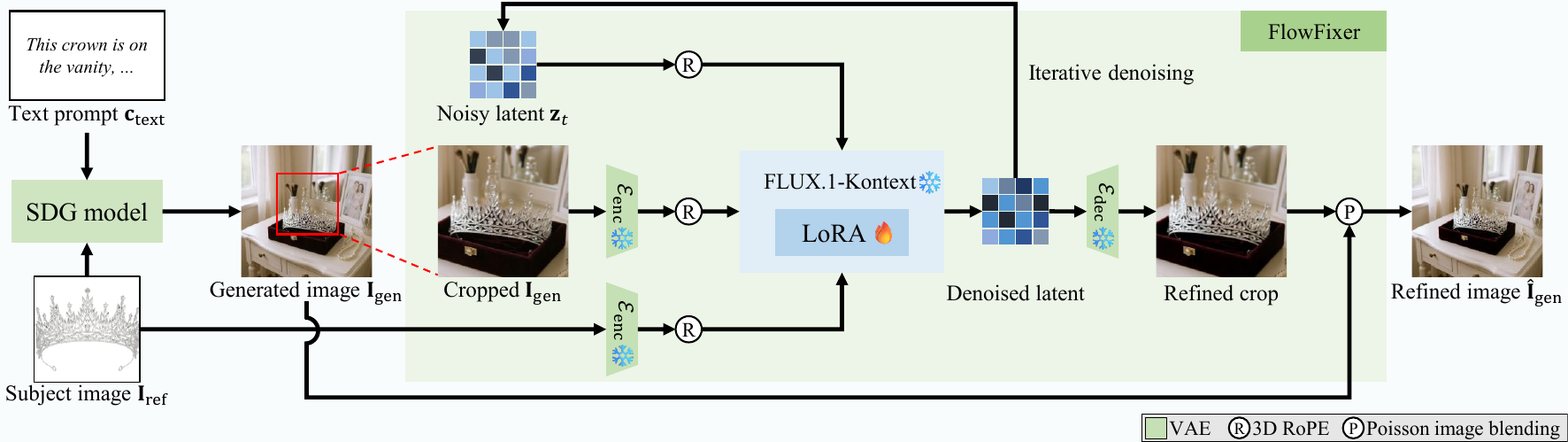}
  \caption{\textbf{FlowFixer inference pipeline.} The model takes two conditional inputs: reference subject image $\bI_{\text{ref}}$ and the generated image $\bI_{\text{gen}}$ from any SDG model. Then the model produces a refined result $\widehat{\bI}_{\text{gen}}$ that preserves global layout. For faster inference, we optionally refine only a subject-centric crop of $\bI_{\text{gen}}$ and blend it back using Poisson image blending.}
  \label{fig:overview}
\end{figure*}

\section{Method}
\label{sec:method}
\subsection{Diffusion preliminaries}
Diffusion models are probabilistic generative frameworks that progressively transform a simple prior $p_s$ (\textit{e.g.}, Gaussian noise) into samples from a target distribution $p_t$ through iterative denoising or continuous flows.
Let $\bx_t$ (or $\bz_t$ in latent diffusion) denote the state at time $t\in[0,1]$ along this trajectory.
The generative process starts from noise and can be formally expressed as
\begin{equation}
\bx_1 \sim p_s, \qquad \bx_0 = \mathcal{D}(\bx_1) \sim p_t,
\end{equation}
where $\mathcal{D}$ represents a learned denoising or flow-matching process~\cite{ho2020denoising, song2020score, lipman2022flow}.
A key property of diffusion models is their conditioning flexibility:
\begin{equation}
\bx_0 = \mathcal{D}(\bx_1, \mathbf{c}),
\label{eq:diffusion_condition}
\end{equation}
where $\mathbf{c}$ denotes auxiliary controls such as text prompts or reference images.
In latent diffusion models~\cite{rombach2022high, flux2024}, $\bx_t$ corresponds to a latent variable $\bz_t$ encoded by a VAE $\mathcal{E}$, and the final image is reconstructed from the latent sample $\bz_1$. Diffusion models have achieved highly realistic and semantically coherent image generation, 
driven by large-scale architectures and training on massive, diverse datasets.

\subsection{Problem formulation}
Subject-driven generation (SDG) can be viewed as a specific instance of conditional diffusion in Eq.~\ref{eq:diffusion_condition}.
Given a subject reference image $\bI_{\text{ref}}$ and a textual description $\mathbf{c}_{\text{text}}$, 
an SDG model $\mathcal{D}_{\text{SDG}}$ generates a novel scenic image $\bI_{\text{gen}}$ conditioned on both inputs:
\begin{equation}
\bI_{\text{gen}} = \mathcal{D}_{\text{SDG}}(\bz_1, \bI_{\text{ref}}, \mathbf{c}_{\text{text}}), \qquad \bz_1 \sim p_s,
\end{equation}
where $\mathbf{c}_{\text{text}}$ provides high-level semantics and $\bI_{\text{ref}}$ encodes subject appearance cues.

While diffusion models achieve strong global realism and semantic consistency, 
text-conditioned variants often prioritize global coherence over local structural fidelity. 
This limitation stems from the ambiguity of textual conditioning, which captures broad semantics but lacks precise visual cues such as small textures or logos. 
As a result, diffusion models tend to favor perceptual plausibility at the expense of subject-specific details~\cite{zhou2025fireedit,mou2024diffeditor,guo2024focus,thakral2025fine,qiao2024baret}. 
Despite notable progress in large-scale foundation models—%
including Qwen~\cite{bai2023qwen},  FLUX.1-Kontext~\cite{labs2025flux}, and Nano Banana~\cite{comanici2025gemini}—%
fine-grained subject fidelity remains a persistent challenge.

To address this, we design a text-free diffusion-based refiner $\mathcal{D}_\text{refine}$ 
that re-generates $\bI_{\text{gen}}$ under the guidance of $\bI_{\text{ref}}$ through a conditional diffusion process 
starting from latent noise $\bz_1 \sim p_s$ as follows,
\begin{equation}
\widehat{\bI}_{\text{gen}} = \mathcal{D}_\text{refine}(\bz_1, \bI_{\text{gen}}, \bI_{\text{ref}}).
\label{eq:objective}
\end{equation}
Here, $\widehat{\bI}_{\text{gen}}$ preserves the global layout of $\bI_{\text{gen}}$ 
while restoring subject-consistent details from $\bI_{\text{ref}}$. 
Unlike conventional inpainting methods that rely on explicit masks or user interaction, Eq.~\ref{eq:objective} denotes fully automatic refinement without external inputs. 
Furthermore, $\mathcal{D}_\text{refine}$ is optimized with self-supervised pseudo pairs, 
enabling scalable and annotation-free enhancement beyond mask-based editing.
We refer to this refiner as \textbf{FlowFixer}, reflecting its ability to restore fine structural consistency by correcting disrupted feature flow between $\bI_{\text{gen}}$ and $\bI_{\text{ref}}$.

\begin{figure}
  \centering
  \includegraphics[width=0.93\columnwidth]{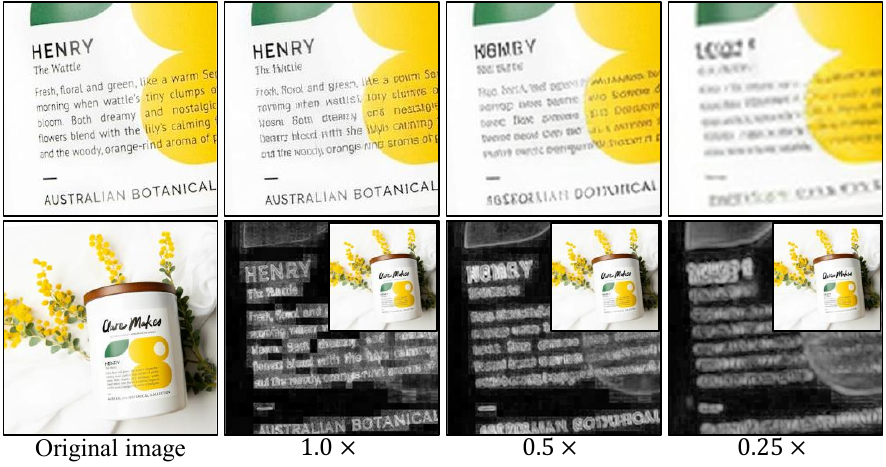}
  \caption{\textbf{Example of one-step denoising distortions.} For each distortion level, pixel-wise variance maps are computed over 10 degraded samples. Insets show example outputs, with distortions concentrated in high-frequency regions.}
  \label{fig:distortion_example}
  \vspace{-3mm}
\end{figure}

\subsection{Pseudo-paired training data}
\label{subsec:pseudo_paired_training_data}
A key challenge in training $\mathcal{D}_\text{refine}$ is the lack of paired data where only subject details are degraded while global structure remains unchanged.
To address this, we construct pseudo pairs $(\bI_{\text{degraded}}, \bI_{\text{clean}})$ from real images by mocking SDG's degradation using a one-step denoising process as follows:
\begin{enumerate}
\itemsep0.2em
\item Start from a clean real image $\bI_{\text{clean}}$.
\item Add noise to $\bI_{\text{clean}}$ and apply a single-step denoising using an off-the-shelf diffusion model~\cite{podell2023sdxl}.
\item Control the degradation level by downscaling $\bI_{\text{clean}}$ to $1.0\times$, $0.5\times$, or $0.25\times$ its original resolution before VAE encoding.
\end{enumerate}
To verify that this process mainly affects fine details, we generate 10 variants with different random seeds in step~2 and compute per-pixel variance maps across them. As shown in Figure~\ref{fig:distortion_example}, the variance concentrates in high-frequency regions while remaining low in smooth backgrounds, confirming that the perturbation minimally disturbs global structure. For step~2, we use SDXL~\cite{podell2023sdxl}.

During training, we treat $\bI_{\text{degraded}}$ as the generated input $\bI_{\text{gen}}$ in Eq.~\ref{eq:objective}. The reference $\bI_{\text{ref}}$ is a spatially perturbed version of the clean image $\bI_{\text{clean}}$, created through random cropping, rotation, or mild color augmentation—and vice versa for data diversity. This setup enables $\mathcal{D}_\text{refine}$ to focus on recovering fine details by learning local correspondences between $\bI_{\text{degraded}}$ and $\bI_{\text{ref}}$, without depending on strict pixel-wise alignment.

\subsection{Training pipeline}
\paragraph{Network architecture.}
FlowFixer builds on FLUX.1-Kontext~\cite{labs2025flux} to leverage image-native editing capability. For a text-free pipeline, we discard the original text token and introduce an additional image input, as illustrated in Figure~\ref{fig:overview}. Consequently, FlowFixer takes three inputs, $\bz_1$, $\bI_{\text{gen}}$, and $\bI_{\text{ref}}$. The images $\bI_{\text{gen}}$ and $\bI_{\text{ref}}$ are encoded by the pretrained VAE into latent tokens, which are concatenated with $\bz_1$ before being processed by the DiT backbone. We adopt 3D RoPE with per-stream timestep offsets ($0$ for $\bz_1$, $1$ for $\bI_{\text{gen}}$, $2$ for $\bI_{\text{ref}}$), to maintain stream separation while allowing full cross-attention.

To discover dense correspondences between $\bI_{\text{gen}}$ and $\bI_{\text{ref}}$, we adopt an explicit dual-stream conditioning mechanism operating in a shared spatial space. This design enforces alignment between the two inputs and facilitates localized refinements guided by the reference. The alignment is further strengthened by our self-supervised, pseudo-paired training scheme.

\vspace{-3mm}
\paragraph{Implementation details.}
We fine-tune FLUX.1-Kontext using LoRA~\cite{hu2022lora} with rank 192, specializing the model for automatic refinement while keeping the parameter overhead minimal.
Training is conducted for 50K iterations with a batch size of 4.
For each iteration, a pseudo training pair $(\bI_{\text{degraded}}, \bI_{\text{clean}})$ is sampled as described in Sec.~\ref{subsec:pseudo_paired_training_data}.
One degraded variant is randomly selected from the three downscaling levels ($1.0\times$, $0.5\times$, $0.25\times$) to ensure balanced degradation diversity.
The model is trained using a mean squared error (MSE) loss between the refined output and the clean target. We use a guidance scale of 1.0 during training and 2.5 at inference, respectively.

We use 18,450 high-quality real-world photographs from Unsplash~\cite{unsplash} to construct the pseudo pairs for training.
The images span diverse objects, materials, and lighting conditions, providing sufficient visual diversity for self-supervised refinement.

\subsection{Crop-based refinement}
While high-resolution generation is critical for subject fidelity, a full-resolution global pass incurs substantial latency and memory cost.
Instead, FlowFixer preserves the background layout while selectively restoring subject details, enabling crop-based refinement during inference, as illustrated in Figure~\ref{fig:overview}.
We first determine a subject-centric crop using keypoint matching~\cite{jiang2024omniglue} between a subject and its generated image, and then refine only this region and paste the result back into the original image.
Since the global structure remains unchanged and only fine details are corrected, simple image-domain blending (\textit{e.g.}, Poisson blending) achieves seamless integration without user-defined masks or inversion.
This substantially reduces runtime and memory while retaining subject-level fidelity.

\begin{table*}[!t]
    \scriptsize
    \setlength{\tabcolsep}{5pt}
    \centering
    \vspace{-3mm}
    \caption{\textbf{Refinement performance on the \FullDatasetName{}.} For all metrics, higher numbers indicate better performance.}
    \label{tb:SDG_refine_performance_full}
    \vspace{-3mm}
    \begin{tabular}{l|cccc|cccc|cccc}
    \toprule
    \multirow{2}{*}{Method} & \multicolumn{4}{c|}{FLUX.1-Kontext-Pro} & \multicolumn{4}{c|}{Qwen-Image-Edit} & \multicolumn{4}{c}{Nano-Banana-Edit}\\
     & AKI $\uparrow$ & $\KGain$ $\uparrow$ & CLIP-I $\uparrow$ & DINO $\uparrow$
     & AKI $\uparrow$ & $\KGain$ $\uparrow$ & CLIP-I $\uparrow$ & DINO $\uparrow$
     & AKI $\uparrow$ & $\KGain$ $\uparrow$ & CLIP-I $\uparrow$ & DINO $\uparrow$\\
    \toprule
    Baseline & - & - & 0.776 & 0.663 & - & - & \textbf{0.777} & \textbf{0.668} & - & - & \textbf{0.796} & \textbf{0.711}\\
    \midrule
    Text-based editing~\cite{labs2025flux}
    & 7.5 & 52.7\% & 0.763 & 0.647
    & 11.1 & 54.1\% & 0.762 & 0.647
    & 61.7 & 77.0\% & 0.782 & 0.691 \\
    OminiControl~\cite{tan2025ominicontrol} + F-Dev
    & 29.0 & 53.9\% & 0.724 & 0.551 
    & 0.48 & 43.8\% & 0.724 & 0.552 
    & \textbf{108.0} & 70.7\% & 0.747 & 0.605 \\
    OminiControl~\cite{tan2025ominicontrol} + F-Kontext
    & 0.49 & 45.7\% & 0.766 & 0.649 
    & -3.37 & 46.8\% & 0.765 & 0.649 
    & 53.0 & 56.6\% & 0.786 & 0.696 \\
    FlowFixer (ours)
    & \textbf{66.5} & \textbf{77.9\%} & \textbf{0.778} & \textbf{0.668} 
    & \textbf{54.0} & \textbf{74.8\%} & \textbf{0.777} & \textbf{0.668} 
    & 64.7 & \textbf{79.2\%} & \textbf{0.796} & \textbf{0.711} \\
    \bottomrule
    \end{tabular}
\end{table*}

\begin{figure*}[!t]
  \centering
  \includegraphics[width=0.93\textwidth]{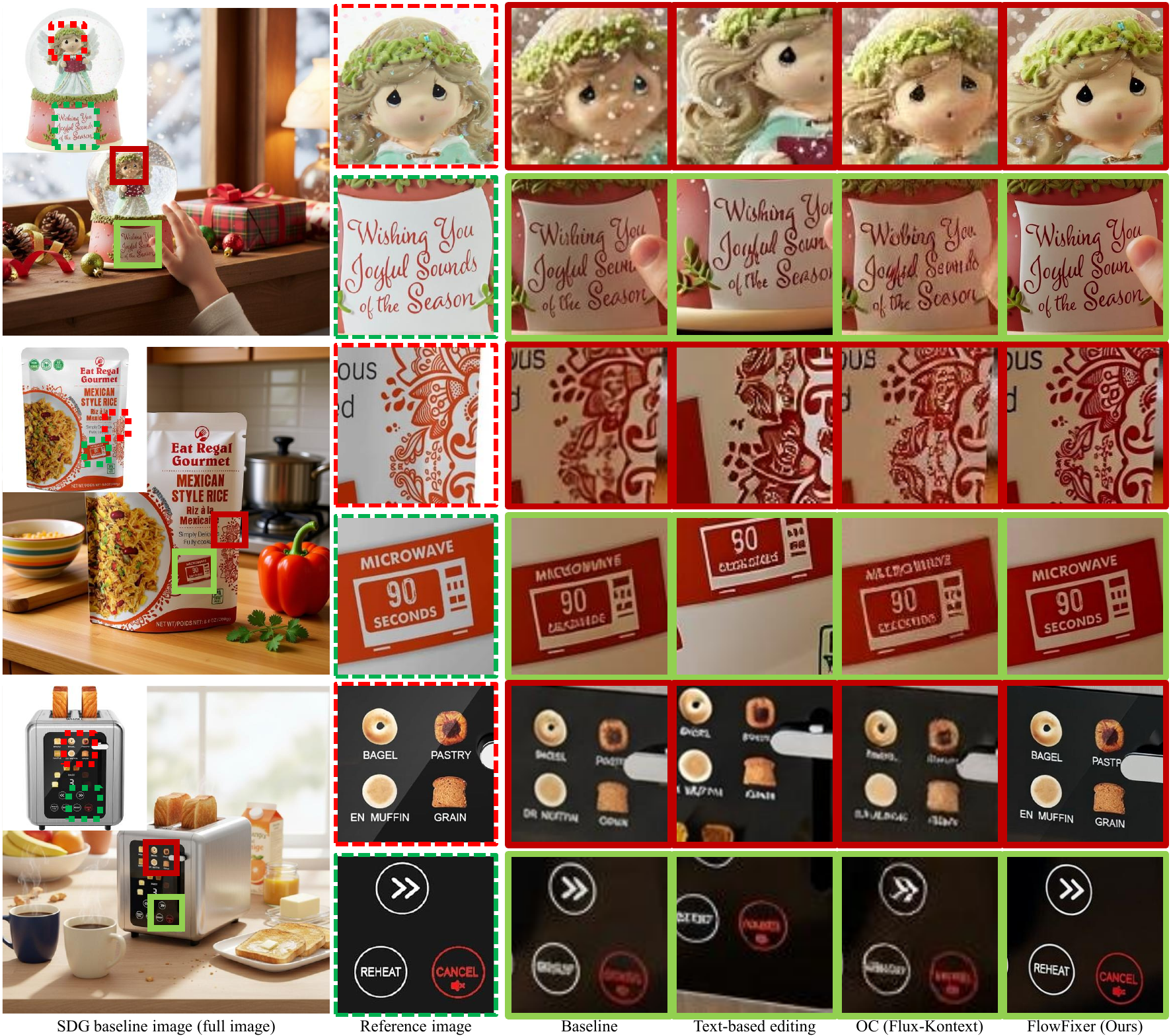}
  \caption{\textbf{Qualitative comparison on Subject fidelity refinement on the \FullDatasetName{} dataset.} The insets in the full images show the reference subject images and the red and green boxes indicate the zoomed-in regions. The regions for zoomed-in views are found on the SDG baseline images and cropped the same area for all methods.}
  \label{fig:main}
  \vspace{-5mm}
\end{figure*}

%% file: src/detail_aware_evaluation.tex
\section{Detail-aware Evaluation}
\label{sec:detail_aware_evaluation}

\subsection{Evaluation metric}
While widely used, existing perceptual metrics fall short in evaluating fine-grained details.
Common similarity measures, such as CLIP~\cite{radford2021learning} or DINOv2~\cite{oquab2023dinov2} primarily capture global semantics and overlook high-frequency fidelity, making them unsuitable for assessing detail consistency.

To better quantify subject fidelity, we exploit keypoint matching that finds dense correspondences between the reference and generated images. This approach is based on the observation that images with better subject fidelity yield a higher number of matched keypoints.
We define two metrics: i) absolute keypoint increase (AKI) and ii) keypoint matching gain $\KGain$. First, we formulate AKI by
\begin{equation}
\text{AKI} =
\mathcal{N}(\KeypointModel{}(\bI_{\text{ref}}, \widehat{\bI}_{\text{gen}})) - \mathcal{N}(\KeypointModel{}( \bI_{\text{ref}}, \bI_{\text{gen}})),
\end{equation}
where $\mathcal{N}(\KeypointModel{}(a, b))$ denotes the number of matched keypoints between $a$ and $b$ using the keypoint matching network $\KeypointModel{}$. A higher AKI score indicates stronger preservation of subject-specific fine details and structural alignment.

While AKI effectively quantifies instance-level improvements, its absolute values depend on the choice and calibration of the keypoint matcher, and thus are not strictly comparable across settings. Moreover, when averaged over a large set, many small yet consistent improvements can be diluted by a few large changes, obscuring the overall trend. Therefore, we also calculate the keypoint matching gain, $\KGain$, which averages the fraction of cases that improve. Formally, we define
\begin{equation}
\KGain
= \frac{1}{N}\sum_{i=1}^{N}
\delta(\text{AKI}_i, \tau)
\end{equation}
where $\delta$ is a binary indicator function that becomes 1 when $\text{AKI}_i$ is higher than $\tau$ and 0 otherwise. $\text{AKI}_i$ is an AKI score of $i$-th image sample in a dataset. We report $\KGain$ in percentage and set $\tau{=}0$ as default. These metrics effectively capture enhancement of local fidelity. For evaluation, we employ an off-the-shelf keypoint matching network, OmniGlue~\cite{jiang2024omniglue}.

\vspace{3mm}
\subsection{Evaluation dataset}
Existing SDG benchmarks~\cite{ruiz2023dreambooth, peng2024dreambench} primarily focus on global realism and semantic alignment rather than preserving fine-grained subject fidelity.
As a result, their subject categories are often visually simple and contain limited texture or detail (\textit{e.g.}, rubber ducks, plants, or cartoon figures).
To achieve rigorous evaluation of subject fidelity, we introduce \textbf{\FullDatasetName{}}, a large-scale, subject-diverse benchmark structured by subject–prompt pairs.
To construct the dataset, we first collected 29K subject images and generated prompts for SDG using a vision-language model (VLM), Claude 3.5. 
For each prompt-subject pair, we generated five variants per SDG baseline. We used three SDG baselines (FLUX.1-Kontext-Pro~\cite{labs2025flux}, Qwen-Image-Edit~\cite{bai2023qwen}, and Nano Banana-Edit~\cite{comanici2025gemini}), which led to 435K SDG images in total. 
Finally, we applied a coarse quality filter to ensure that the subject is clearly present in the SDG image. After the filtering, the final dataset consists of 258K subject - SDG image pairs.

For controlled and reproducible studies, we also curate a fixed subset, \textbf{\SmallDatasetName{}} from the \FullDatasetName{} dataset, collecting 100 images from each backbone. \SmallDatasetName{} preserves the distribution of baseline match counts while balancing categories, and we reuse this fixed subset for all ablations and human evaluations to ensure comparability and reproducibility. 



%% file: src/experiments.tex
\section{Experiments}
\label{sec:experiments}
\subsection{Subject fidelity refinement}
\label{subsec:subject_fidelity}
Table~\ref{tb:SDG_refine_performance_full} reports refinement results on \FullDatasetName{} dataset under the three SDG baselines (\textit{i.e.}, FLUX.1-Kontext-Pro, Qwen-Image-Edit, and Nano-Banana-Edit). In Table~\ref{tb:SDG_refine_performance_full}, we compare four different refinement models, including the proposed FlowFixer. The compared refinement models are:

\begin{itemize}\itemsep0.5em
\item \textbf{Text-based editing:} FLUX.1-Kontext, which is a text-based editing model, accepting the subject and SDG images concatenated on the x-axis while refinement is guided by an input prompt.
\item \textbf{OminiControl + FLUX.1 (Dev/Kontext):} OminiControl~\cite{tan2025ominicontrol} fine-tuned on FLUX.1-Dev and FLUX.1-Kontext, respectively, using our pseudo-paired dataset. We use the same training data as FlowFixer.
\end{itemize}


\noindent Since there is no algorithm tailored for refinement of SDG, we finetuned OminiControl~\cite{tan2025ominicontrol} with state-of-the-art backbones~\cite{flux2024, labs2025flux}, which can accept a subject as a conditional input, and used the text-based editing method~\cite{labs2025flux} for benchmarking.

Note that AKI and $\KGain$ are not obtainable for the baselines since these metrics quantify the changes relative to the baseline's SDG output. We additionally report CLIP-Image (CLIP-I) and DINOv2 similarity as complementary perceptual metrics. To isolate subject fidelity from background content, similarities are computed only on the subject regions by detecting the bounding box of the subject. 

We summarize our statistical and empirical findings from Table~\ref{tb:SDG_refine_performance_full} and Figure~\ref{fig:main} as follows:

\begin{enumerate}[label=(\roman*),leftmargin=10pt,itemsep=2pt]
\item As illustrated in Figure~\ref{fig:main}, FlowFixer restores fine details of the subject while preserving the original image layout. In contrast, the other refinement models either shift the scene or fail to improve local structure. For example, `Text-based editing' often maintains semantics but alters composition, undermining subject consistency. In contrast, FlowFixer avoids such layout drift while increasing local correspondences.

\item Quantitatively, across all SDG backbones, FlowFixer consistently outperforms its alternatives in AKI and achieves an average $\KGain$ of 77.3\%, demonstrating model-agnostic robustness.

\item Interestingly, these keypoint-based gains are not reflected in CLIP-I or DINOv2 scores, which remain nearly unchanged. This indicates that common perceptual metrics overlook fine-grained structural fidelity, reinforcing the need for specialized metrics like AKI and $\KGain$.

\item While alternative fine-tuned models (OminiControl + FLUX.1-Dev and Kontext) occasionally increase AKI, their $\KGain$ often drops below 50\%, meaning such methods show no consistent pattern of improvement.

\item On Nano Banana, certain methods achieve inflated keypoint metrics by copy-pasting the subject or synthesizing a new scene with larger subject rather than refining the given output. This results in the elevated AKI scores but reduced global consistency, as reflected in the lower CLIP-I and DINOv2 similarities on cropped subject regions. 
\end{enumerate}

\begin{figure}[!t]
  \centering
  \includegraphics[width=\linewidth]{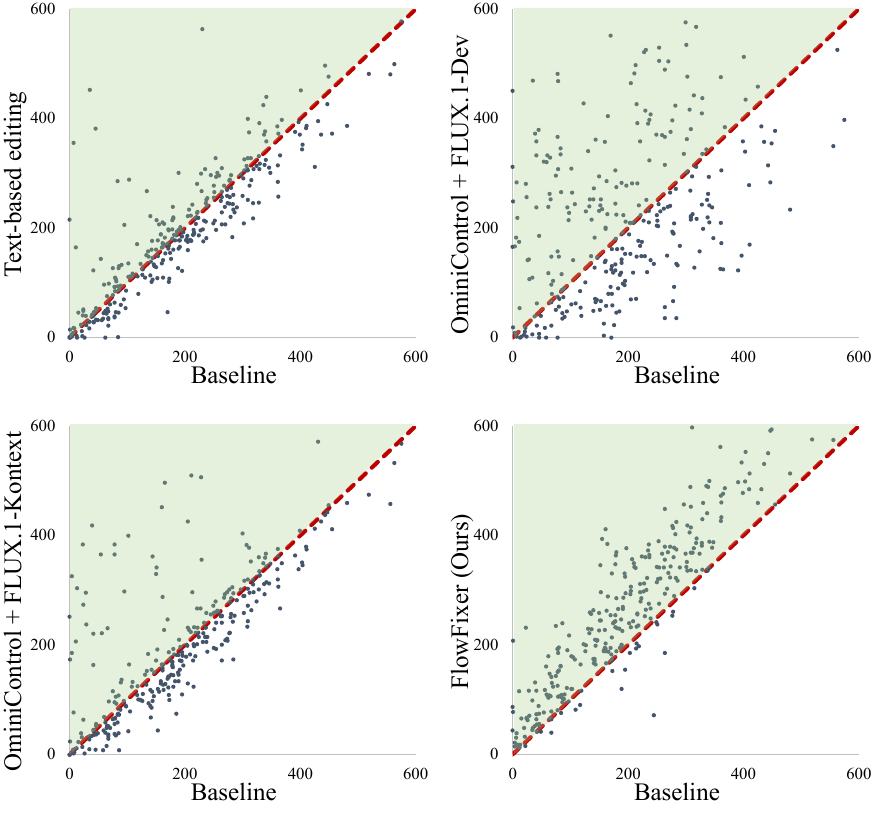}
  \caption{\textbf{Scatter plots of the number of keypoint matches on \SmallDatasetName{}.} Each dot represents a sample; points above the red dashed line indicate an increase in keypoint matches (positive AKI, green region), suggesting improved structural alignment and subject fidelity. Samples below the line show decreased correspondence after refinement.}
  \label{fig:scatter_plot}
\end{figure}

Figure~\ref{fig:scatter_plot} shows scatter plots of keypoint changes before and after refinement on \SmallDatasetName{}. Among all methods, only FlowFixer reveals a consistent and directional pattern, reliably increasing the number of matched keypoints (AKI) across most samples. In contrast, alternative methods exhibit no clear trend, with improvements occurring sporadically and often accompanied by regressions. This further highlights the robustness and generalizability of FlowFixer's refinements.


\begin{table}[!t]
    \footnotesize
    \setlength{\tabcolsep}{8pt}
    \centering
    \vspace{-3mm}
    \caption{\textbf{Refinement performance compared to original SDG images on the \SmallDatasetName{}.} For all metrics, higher numbers indicate better performance.}
    \label{tb:SDG_refine_performance_subset}
    \vspace{-3mm}
    \begin{tabular}{l|ccc}
    \toprule
    Method & AKI $\uparrow$ & $\KGain$ $\uparrow$ & VLM $\uparrow$\\
    \toprule
    Text-based editing~\cite{labs2025flux} & 1.87 & 45.9\% & 41.3\% \\
    OminiControl~\cite{tan2025ominicontrol} + F-Dev & 22.7 & 46.6\% & 4.2\%\\
    OminiControl~\cite{tan2025ominicontrol} + F-Kontext & 11.1 & 38.4\% & 25.2\%\\
    FlowFixer (ours) & \textbf{67.3} & \textbf{91.2\%} & \textbf{79.0\%}\\
    \bottomrule
    \end{tabular}
\end{table}

\begin{figure}[!t]
  \centering
  \includegraphics[width=\linewidth]{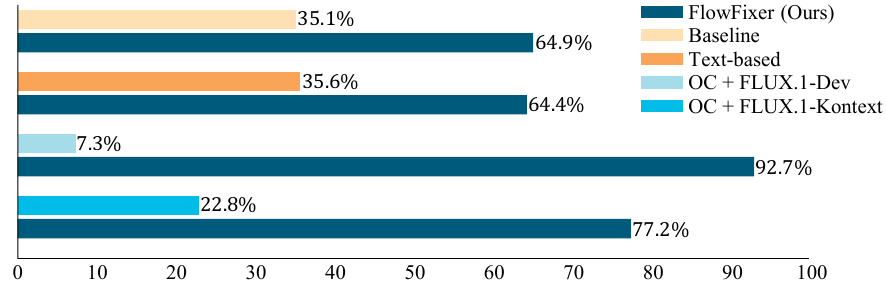}
  \caption{\textbf{A/B study results comparing the FlowFixer against four alternatives.} FlowFixer is consistently preferred by human raters.}
  \label{fig:human_eval}
\end{figure}

\subsection{Human Evaluation and VLM Judgment}
\label{subsec:human_alignment}
To assess how well our metrics reflect human perception, we conducted an A/B tests on the \SmallDatasetName{} subset using Amazon Mechanical Turk. For each test case, human evaluators were shown the reference subject alongside two candidate images, \textit{i.e.}, FlowFixer vs.\ one alternative method. Then, we asked the evaluators to choose the one that better preserves subject-specific details. Each pair was evaluated by five independent evaluators, and responses were aggregated across the dataset. 

Figure~\ref{fig:human_eval} shows that the human evaluators strongly prefer FlowFixer over the baseline and the other refinement methods, which align with our proposed metrics, AKI and $\KGain$. Notably, FlowFixer's advantages over the baseline and the text-based editing~\cite{labs2025flux} are comparable (64.9\% and 64.4\%), suggesting that a text prompt for editing only makes a negligible difference in terms of subject fidelity. Moreover, FlowFixer is favored over OminiControl variants~\cite{tan2025ominicontrol} by even larger margins (92.7\% and 77.2\%).

In addition to the human evaluation, we also evaluate metric agreement with a Vision-Language Model (VLM), Claude 3.7, serving as an automated judge. For each case, the VLM receives the reference image and two subject-region crops (Baseline vs.\ one alternative). To mitigate order bias, we present two images in both A/B and B/A orders and average the decisions.

As shown in Table~\ref{tb:SDG_refine_performance_subset}, VLM judges FlowFixer to be the best restoration method in terms of subject fidelity. In addition to that, VLM judgments exhibit strong alignment with AKI and $\KGain$, cross-validating their effectiveness in capturing perceptual improvements in subject fidelity.

\begin{figure}[!t]
  \centering
  \includegraphics[width=\linewidth]{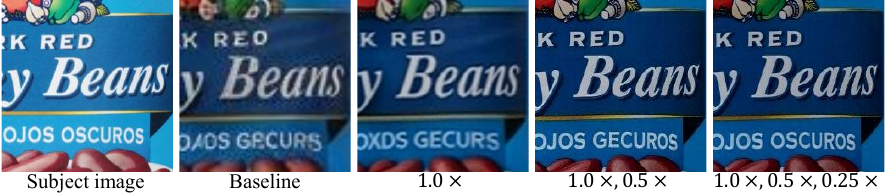}
  \caption{\textbf{Impact of training distortion levels on refinement performance.} Using a range of distortion levels during training enhances the model's ability to handle diverse degradation at inference time, resulting in more robust restoration.}
  \label{fig:distortion_levels_for_training}
\end{figure}

\begin{figure}[!t]
  \centering
  \includegraphics[width=\linewidth]{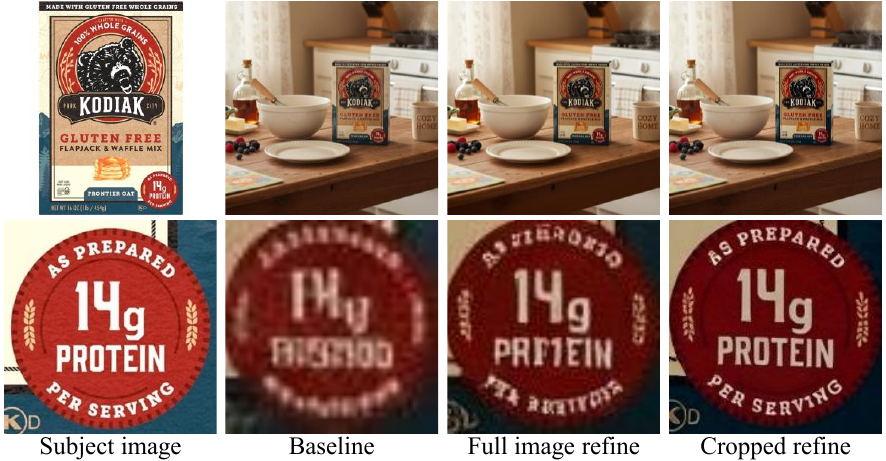}
  \caption{\textbf{Efficacy of cropped refinement in comparison with full image refinement.} While full image refinement moderately enhances subject fidelity, cropping further improves legibility.}
  \label{fig:crop_based_refine}
\end{figure}

\subsection{Distortion levels for training}
To assess the impact of degradation diversity during training, we compare FlowFixer models trained with different subsets of distortion levels: (i) only slight noise ($1.0\times$), (ii) moderate and slight noise ($1.0\times$, $0.5\times$), and (iii) the full range ($1.0\times$, $0.5\times$, $0.25\times$). As illustrated in Figure~\ref{fig:distortion_levels_for_training}, including various levels of distortion during training significantly boosts robustness, especially under large-scale artifacts, highlighting the importance of diverse degradation simulation for effective refinement.


\subsection{Crop-based refinement}
Figure~\ref{fig:crop_based_refine} compares a single global refinement pass against our crop-based refinement strategy. In both cases, the global scene composition remains unchanged, highlighting FlowFixer's inherent stability with respect to layout drift. Notably, even under the same evaluation resolution, crop-based refinement yields more accurate recovery of fine-grained subject details, thanks to its focused and localized processing. This allows better fidelity in details without compromising global coherence.

%% file: src/conclusion.tex
\section{Conclusion}
We introduced FlowFixer, a model-agnostic detail refiner for subject-driven generation that recovers fine structural details while preserving global layout. Trained on self-supervised pseudo pairs simulating high-frequency degradation, FlowFixer scales to in-the-wild references without paired subject–scene data. Our text-free, direct image-to-image formulation avoid prompt ambiguity and consistently improve fidelity. Paired with keypoint-matching-based metrics for ground-truth-free evaluation, FlowFixer demonstrates superior performance across diverse SDG methods. Future directions include (i) multi-reference refinement that leverages multiple reference images, and (ii) user-interactive correction using auxiliary control signals, such as scribble masks.
